\documentclass[10pt,twocolumn,letterpaper]{article}

\usepackage{cvpr}
\usepackage{times}
\usepackage{epsfig}
\usepackage{graphicx}
\usepackage{amsmath}
\usepackage{amssymb}
\usepackage{enumitem}
\usepackage{mathrsfs}
\usepackage{algorithm}
\usepackage{algorithmic}
\usepackage{setspace}
\usepackage{subfigure}

\usepackage[breaklinks=true,bookmarks=false]{hyperref}

\cvprfinalcopy 

\def\cvprPaperID{8784} 

\begin{document}

\title{State-Relabeling Adversarial Active Learning}

\author{
	Beichen Zhang\textsuperscript{\rm 1},
	Liang Li\textsuperscript{\rm 2}\thanks{Corresponding author},
	Shijie Yang\textsuperscript{\rm 1, 2},
	Shuhui Wang\textsuperscript{\rm 2},
	Zheng{-}Jun Zha\textsuperscript{\rm 3},
	Qingming Huang\textsuperscript{\rm 1, 2, 4}\\
	\textsuperscript{\rm 1}University of Chinese Academy of Sciences, Beijing, China\\
	\textsuperscript{\rm 2}Key Lab of Intell. Info. Process., Inst. of Comput. Tech., Chinese Academy of Sciences, China\\
	\textsuperscript{\rm 3}University of Science and Technology of China, China, $^4$Peng Cheng Laboratory, Shenzhen, China, \\
	{\tt\small 
	\{beichen.zhang, shijie.yang\}@vipl.ict.ac.cn, 
	\{liang.li, wangshuhui\}@ict.ac.cn, 
	}\\
	{\tt\small
	zhazj@ustc.edu.cn,
	qmhuang@ucas.ac.cn}
}

\maketitle
\thispagestyle{empty}

\begin{abstract}
Active learning is to design label-efficient algorithms by sampling the most representative samples to be labeled by an oracle.
In this paper, we propose a state relabeling adversarial active learning model (SRAAL), that leverages both the annotation and the labeled/unlabeled state information for deriving the most informative unlabeled samples. The SRAAL consists of a representation generator and a state discriminator. The generator uses the complementary annotation information with traditional reconstruction information to generate the unified representation of samples, which embeds the semantic into the whole data representation. Then, we design an online uncertainty indicator in the discriminator, which endues unlabeled samples with different importance. As a result, we can select the most informative samples based on the discriminator's predicted state. We also design an algorithm to initialize the labeled pool, which makes subsequent sampling more efficient. The experiments conducted on various datasets show that our model outperforms the previous state-of-art active learning methods and our initially sampling algorithm achieves better performance.
\end{abstract}


\section{Introduction}

\begin{figure}[t]
\begin{center}

\includegraphics[width=0.9\linewidth,trim=50 150 80 400]{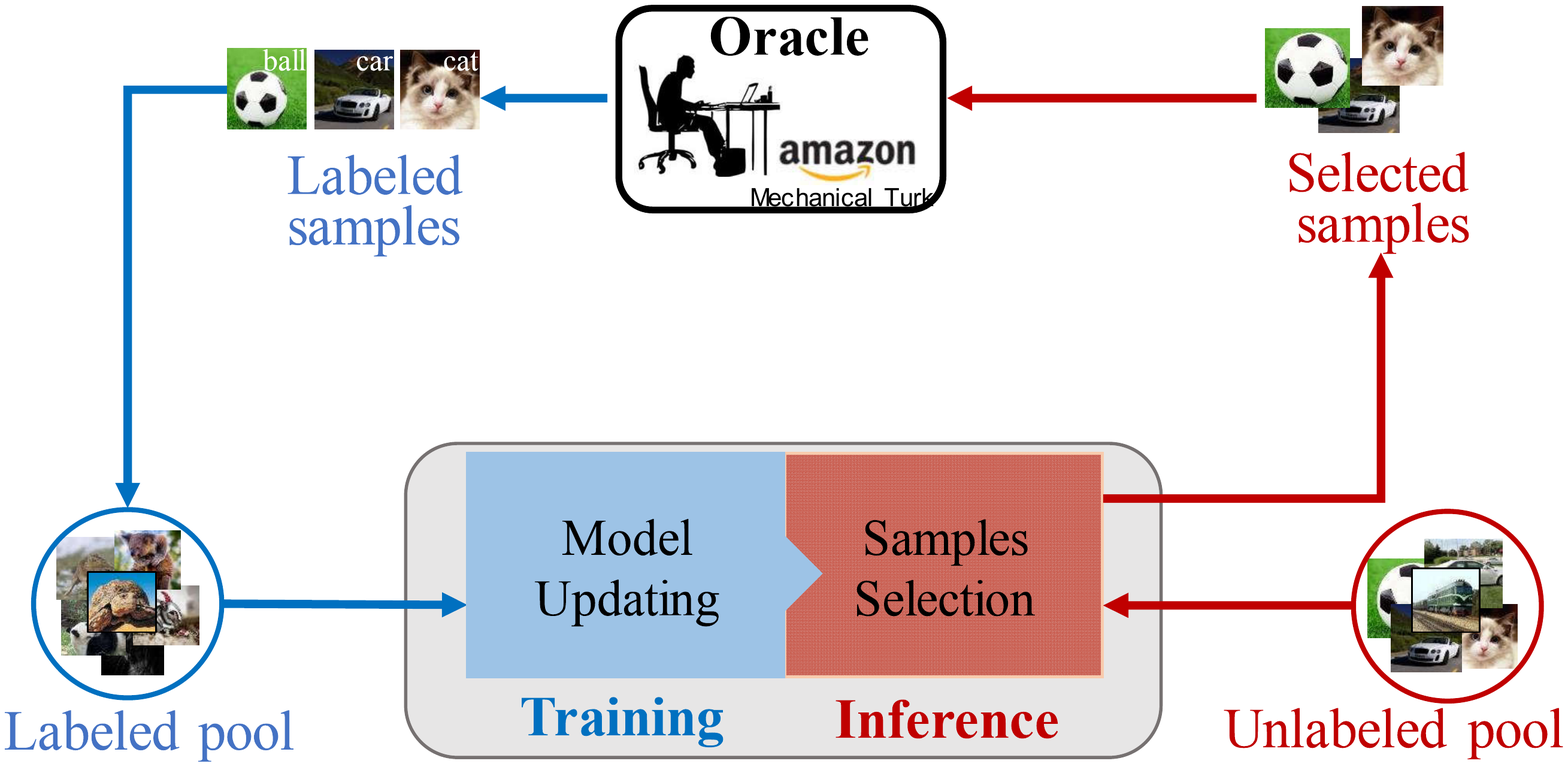}
{\rule{0pt}{2in} \rule{1\linewidth}{0pt}}
\end{center}
   \caption{A traditional pool-based active learning cycle. At each iteration, the sampling model is trained with labeled data. After training, a subset of unlabeled samples is selected based on the model inference and then labeled by an oracle. The active learning system will repeat this iteration until the model performance meets user's requirements or the label budget runs out.}
\label{img1}
\end{figure}

Although deep neural network models have made great success in many areas, they still heavily rely on large-scale labeled data to train large number of parameters. Unfortunately, it is very difficult, time-consuming, or expensive to obtain labeled samples, which becomes the main bottleneck for deep learning methods~\cite{ebrahimi2017gradient}. To reduce the demand of labeled data, learning methods like unsupervised learning~\cite{doersch2015unsupervised,noroozi2017representation}, semi-supervised learning ~\cite{joulin2016learning, mahajan2018exploring}, weakly supervised learning~\cite{ZhouA, Xuejing2019} and active learning have attracted a lot of attention. 
Unsupervised and semi-supervised methods aim to fully utilize the unlabeled samples while active learning is to select as few samples to be labeled as possible for efficient training. 
This paper focuses on active learning, which is widely used in computer vision tasks such as classification~\cite{sener2017active,beluch2018power} and segmentation~\cite{yang2017suggestive,gorriz2017cost}.


In active learning, this means the selected samples should be the most informative ones.
As shown in Fig.~\ref{img1}, active learning algorithm is typically an iterative process in which a set of samples is selected to be labeled from an unlabeled pool at each iteration. These selected unlabeled subsets are labeled by an oracle, integrated into the labeled data pool. How to select the most informative samples from the unlabeled pool is the key problem in active learning.

To solve this problem, previous works have made full use of the annotation information of labeled data from the oracle. Many methods ~\cite{beluch2018power, sener2017active, gal2016dropout,yang2017suggestive} built deep learning models, trained them under the supervision of the labeled samples, and then implemented the inference for sampling. A recent work ~\cite{yoo2019active} (LL4AL) designed a deep network with an auxiliary loss prediction for labeled data, then selected samples based on the predicted loss. 
As the supervised information for network training, both the size and quality of labeled instances determines the performance of models. 
Further, the above approaches usually require a certain amount of labeled samples to achieve a high accuracy. However, in the early iterations of sampling, the labeled pool is usually small, so that it restricts the ability to choose samples with high quality. 


Beside the above annotation information, some recent works focused on utilizing the state information of samples which indicate a sample is labeled (0) or unlabeled (1). 
As the key of active learning is to select unlabeled samples and label them, this state information can be directly used as the supervised information in this process.
Some recent works ~\cite{ducoffe2018adversarial,sinha2019variational} regarded the state information as a kind of adversarial label. They built a discriminator to map the data point to a binary label which is 1 if the sample is unlabeled and is 0, otherwise. 
The above works consider all the unlabeled samples of the same quality. In fact, different samples in unlabeled pool have different importance for target task, and an unlabeled sample has lower priority to be labeled if it is more similar to samples in labeled pool.
Thus, this state information should be deeply explored to leverage the sample selection.

In this paper, we propose a state relabeling adversarial active learning model (SRAAL) that considers both the annotation and the state information for deriving most informative unlabeled samples. 
Our model consists of a unified representation generator and a labeled/unlabeled state discriminator. 
For the generator, we first build an unsupervised image reconstructor based on VAE architecture to learn the rich representation.
Secondly, we design a supervised target learner to predict annotations for labeled samples, where the annotation information is embedded into the representation. Then we concatenate the above representations. 
For the state discriminator, both labeled state and relabeled unlabeled state are used to optimize the discriminator. We propose an online uncertainty indicator to do the state relabeling for unlabeled data, 
which endues the unlabeled samples with different importance.
Specifically, the indicator calculates the uncertainty score for each unlabeled sample as its new state label. State relabeling helps the discriminator to select more instructive samples.

It is notable that most previous works mainly concentrated on the selection strategy in later iterations while the initialization is usually random. However, the initialization of labeled pool has a large influence for subsequent sample selection and performance of active learning.
Here we introduce the k-center~\cite{TeofiloClustering} approach to initialize the labeled pool, where selected samples is a diverse cover for the dataset under the minimax distance. 

Experiments on four datasets at image classification and segmentation tasks show that our method outperforms previous state-of-the-art methods. Further, we implement the ablation study to evaluate the contribution of online uncertainty indicator, supervised target learner and our initial sampling algorithm in SRAAL.

\vspace{0.1cm}
The main contributions of this paper are summarized as:
\begin{enumerate}[label=(\roman*), itemindent=1em]
\vspace{-0.1cm}
\item This paper proposes a state relabeling adversarial active learning model to select most informative unlabeled samples. An online uncertainty indicator is designed to relabel the state of unlabeled data with different importance.
\vspace{-0.1cm}
\item We build an unsupervised image recontructor and a supervised target learner to generate a unified representation of image, where the annotation information is embedded iteratively.
\vspace{-0.1cm}
\item We propose the initially sampling algorithm based on the k-center approach, which makes subsequent sampling more efficient.

\end{enumerate}

\section{Related work}
Active learning has been widely studied for decades and most of the classical methods can be grouped into three scenarios: membership query synthesis ~\cite{angluin1988queries}, stream-based selective sampling ~\cite{dagan1995committee,krishnamurthy2002algorithms} and pool-based sampling. As the acquirement of abundant unlabeled samples becomes easy, most of recent works ~\cite{gal2016dropout,sener2017active,beluch2018power,yoo2019active,sinha2019variational} focus on the last scenarios. Current active learning methods can be divided into two categories: pool-based approaches and synthesizing approaches.

Instead of querying most informative instances from an unlabeled pool, the synthesizing approaches~\cite{mahapatra2018efficient,mayer2018adversarial,zhu2017generative} use generative models to produce new synthetic samples that are informative for the current model. These methods typically introduce various GAN models ~\cite{goodfellow2014generative,mirza2014conditional} or VAE models~\cite{kingma2013auto,sohn2015learning} into their algorithm to generate informative data with high quality. However, the synthesizing approaches still has some disadvantages to overcome, such as high computational complexity and instability of performance~\cite{zhu2017generative}. For this reason, this paper mainly focuses on research of the pool-based approaches.

The pool-based approaches can be categorized as distribution-based and uncertainty-based methods. The distribution approach chooses data points that will increase the diversity of labeled pool. To do so, the model should extract the representation of the data and calculate the distribution based on it.  Previous works~\cite{Liang2012Learning,Kesun2019,Shijie2019} have provided various method to learning the representation.
Some active learning models~\cite{elhamifar2013convex,yang2015multi} optimize the data selection in a discrete space, and ~\cite{nguyen2004active} clusters the informative data points to be selected. Some works~\cite{bilgic2009link,hasan2015context,mac2014hierarchical} focus on how to map the distance of distributions to the informativeness of a data point. Besides, some works estimate the distribution diversity by observing gradient~\cite{settles2008multiple}, future errors~\cite{roy2001toward} or output changes of trained model~\cite{freytag2014selecting,kading2016active}. Sener et al.~\cite{sener2017active} introduce core-set technique into active learning. This method calculates the core-set distance by intermediate features rather than the task-specific outputs, which makes the method applicable to any task and network. The core-set technique has a good performance on datasets with small number of classes. However, the core-set method performs ineffective when the number of classes is big or the data points are in high-dimensions~\cite{sinha2019variational}.

Uncertainty-based approaches do selection by estimating the uncertainties of samples and sampling top-K data points at each iteration. For Bayesian frameworks, ~\cite{kapoor2007active,roy2001toward} estimate uncertainty by Gaussian processes and ~\cite{ebrahimi2019uncertainty} adopts Bayesian neural networks. 
\cite{Zhengjun2012} propose a novel active learning approach based on the optimum experimental design criteria in statistics.
These traditional methods perform well in some specific tasks but do not scale to deep learning network and large-scale datasets. The ensemble model method was proposed by ~\cite{beluch2018power} and applied to some specific tasks ~\cite{yang2017suggestive}. ~\cite{gal2016dropout} introduces Monte Carlo Dropout to build multiple forward passes, which is a general method for various tasks. However, both the ensemble method and dropout method are computationally inefficient for current deep network and large-scale datasets. Yoo et al. ~\cite{yoo2019active} propose a Learning-Loss method and has shown the state-of-the-art performance. Their model consists of a task module and a loss prediction module that predicts the loss of the task module. The two modules learn together and the target loss of task module is regarded as a ground-truth loss for the loss prediction module. This method only utilizes the annotation information in labeled samples and the loss prediction accuracy is affected by the performance of task module. If the task module is inaccurate, the predicted loss cannot reflect how informative the sample is.

Some recent works combine uncertainty and distribution to select data points using a two-step process. Distribution of data points can represent the labeled or unlabeled pool and uncertainty estimation based on the distribution can be more generalized and accurate. A two-step model calculating uncertainty based on information density was proposed in ~\cite{li2013adaptive}. DFAL~\cite{ducoffe2018adversarial} and VAAL~\cite{sinha2019variational} introduces adversarial learning into their models and build a module to learn the representation of data points. The former method extract representation by learning labeled sample’s annotation, while the VAAL method builds a latent space by a VAE that learns together with the discriminator. Both of these models map the representation to labeled/unlabeled in a brute force way and the labeled/unlabeled information are not equivalent to informativeness. For this reason, the results of this method may be unreliable.
\section{Method}
\subsection{Overview}
In this section, we formally define the scenario of active learning (AL) and set up the notations for the rest of the paper. In the AL, we have a target task and a target model $\Theta$ for the task. At the initial stage, there exists a large unlabeled data pool from which we can randomly select $\mathcal{M}$ samples and obtain annotations of them via an oracle. Let us denote the initial unlabeled pool by $D_U$ and the initial labeled pool by $D_L$. $(x_U)$ denotes that a data point in unlabeled pool and $(x_L, y_L)$ denotes a data point and its annotation in labeled pool.

The key of the AL algorithm is to select the most informative samples from the unlabeled pool $D_U$. Once the labeled and unlabeled data pools are initialized, a fixed number of samples will be iteratively selected, labeled and transferred from the unlabeled pool to the labeled pool. Then the unlabeled and labeled pool are updated. As illustrated in Fig. 1, this procedure will repeat until the performance of the target model meets user's requirements, or the budget for annotation runs out.

\begin{figure*}
\begin{center}
\includegraphics[width=0.9\linewidth,trim=50 190 50 120]{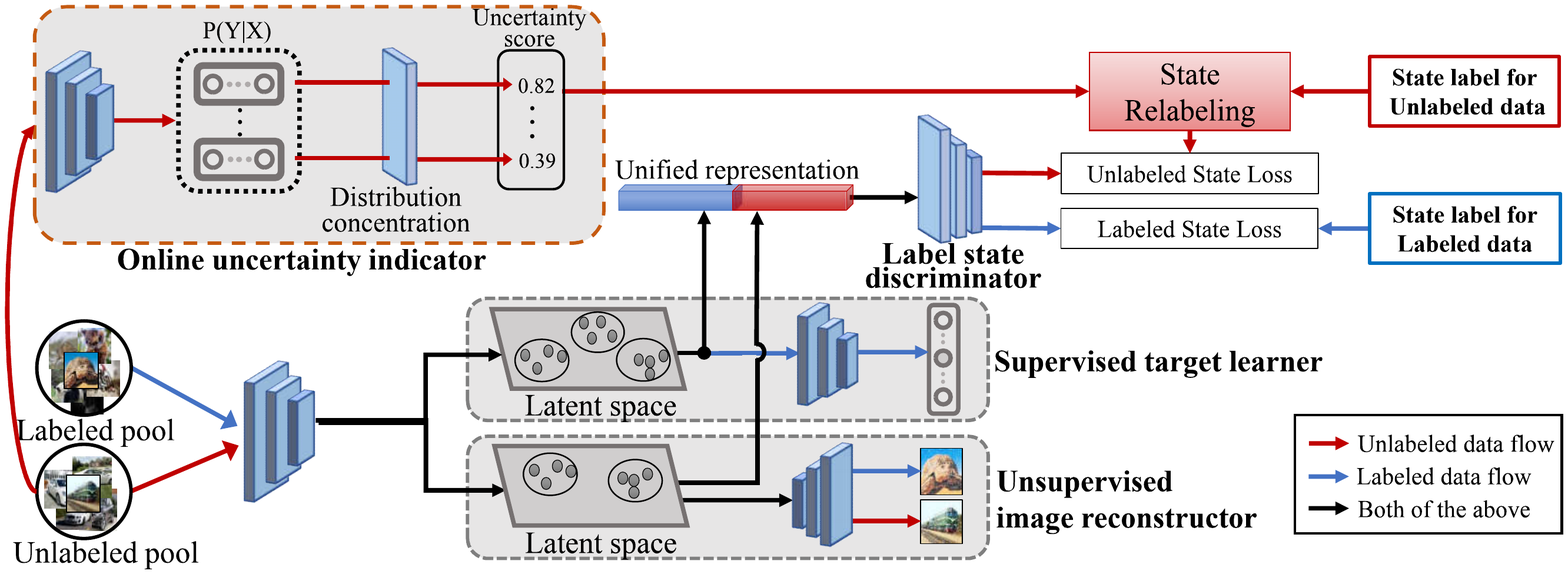}
{\rule{0pt}{0in} \rule{.9\linewidth}{0pt}}
\end{center}
   \caption{Network architecture of our proposed SRAAL. It consists of a unified representation generator and a labeled/unlabeled state discriminator. The generator embeds the annotation information into the final image features via the supervised target learner and unsupervised image reconstructor. Online uncertainty indicator is introduced to relabel the state of unlabeled samples and endues them with different importance. Finally, the state discriminator is updated through the labeled and unlabeled state losses, and helps select the more informative samples.}
\label{img2}
\end{figure*}

Fig.~\ref{img2} shows our state relabeling adversarial active learning model (SRAAL), which uses the annotation and labeled/unlabeled state information for selecting the most informative samples. The SRAAL consists of a unified representation generator (Section~\ref{3.2}) and a labeled/unlabeled state discriminator (Section~\ref{3.3}). 
The former learns the annotation-embedded image feature, and the latter 
selects more representative samples to be labeled with the help of the online uncertainty indicator. Sampling strategy based on the generator and discriminator is introduced in Section~\ref{3.4}, and the proposed initially sampling algorithm with k-center is detailed in Section~\ref{3.5}.

\subsection{Unified representation generator}
\label{3.2}
The image representation learning is in the charge of the unified representation generator which consists of the unsupervised image reconstructor (UIR) and the supervised target learner (STL). The image encoder consists of a CNN and two FC modules. 
The CNN extracts image feature and then FC individually learns the two latent variables for STL and UIR. 
The UIR module is a variational autoencoder (VAE) in which a low dimensional latent space is learned based on a Gaussian prior. As this process does not require annotations and the reconstruction target is the image itself, samples from both the labeled pool and unlabeled pool contribute to this module. As it's a VAE,   the objective function of this module can be formulated as,
\begin{equation}\label{eq1}
\begin{aligned}
&\mathcal{L}^{UIR} = \mathcal{L}_{U}^{UIR} + \mathcal{L}_{L}^{UIR}\\
&\mathcal{L}_{U}^{UIR}=E[log [p_\phi (x_U \mid z_U)] - D_{KL} (q_\theta (z_U \mid x_U) \parallel	p(z))]\\
&\mathcal{L}_{L}^{UIR} = E[log [p_\phi (x_L \mid z_L)] - D_{KL}(q_\theta (z_L \mid x_L) \parallel p(z))]
\end{aligned}
\end{equation}
where $\mathcal{L}_{U}^{UIR}$ is the objective function for unlabeled data points and ${L}_{L}^{UIR}$ is for labeled ones, $z$ is the latent variables, 
$\phi$ parametrizes the decoder $p_\phi$ and $\theta$ parametrizes the encoder $q_\theta$. 
The UIR module finally learns the rich representation by reconstructing both the labeled and unlabeled samples.

To embed the annotation information into the representation, we build a supervised target learner to predict annotations of the samples based on the representation in latent space. The STL is also a VAE network and its decoder does not decode the representation for reconstruction. The decoder of STL varies with different tasks. For example, the decoder is a classifier for image classification, or a segmentation model for semantic segmentation. The STL is similar to VAE but only labeled samples can provide loss for STL's training. We can formulate the objective function for STL as follows,
\begin{equation}\label{eq3}
\mathcal{L}_{L}^{STL} = E[log [p_\phi (y_L \mid z_L)] ] - D_{KL}(q_\theta (z_L \mid x_L) \parallel p(z))]
\end{equation}
where $z_L$ is the latent variables from the latent space for labeled data, 
$\phi$ parametrizes the decoder $p_\phi$ and $\theta$ parametrizes the encoder $q_\theta$.
The STL module will embed the annotation information into the representation. 

The two above  representations are concatenated together as the unified image representation.

\subsection{State discriminator and state relabeling}
\label{3.3}
To make full use of the state information, 
we introduce the adversarial learning into SRAAL, where a discriminator is built to model the state of samples. The previous works utilize the binary state information, where the state of unlabeled samples is set to 1 and that of labeled samples is set to 0. In fact, different samples in unlabeled pool have different contribution for target task, and an unlabeled sample has lower priority to be labeled if it is more similar to samples in labeled pool. 
To better use the state information, we propose the online uncertainty indicator (OUI) to calculate an uncertainty score to relabel the state of unlabeled data. The uncertainty score measures the distribution concentration of the unlabeled data and is bound to [0,1]. After the state relabeling, the state of unlabeled samples changes from the fixed binary label 1 to a new continuous state.


The OUI calculates the uncertainty score based on the prediction vector of the target model(such as image classifier, semantic segmentation model). Before each iteration, the target model is trained with labeled data and then produce a prediction vector for each unlabeled sample. Specifically, for image classification, the prediction is a probability vector for each category. For segmentation, each pixel has a probability vector and the prediction vector is the mean of each probability vector. Assume that the number of classes is $C$ and the samples are labeled with $y_i$ $\in \mathcal{R}^C$.  The calculation of the uncertainty score is formulated as,
\begin{equation}\label{eq4}
Indicator(x_U) = 1 -  \frac{MINVar(V)}{Var(V)}  \times max(V)
\end{equation}
where $x_U$ is an unlabeled sample, $V=p(x_U |D_L)$ is the probability vector of $x_U$ based on the target model trained with current labeled pool $D_L$. 

The $MINVar(V)$ can be formulated as,
\begin{equation}\label{eq41}
\begin{aligned}
&MINVar(V)= Var(V') \\
&=\frac{1}{C}((\frac{1}{C}-max(V))^{2} + (C-1)(\frac{1}{C}-\frac{1-max(V)}{1-C})^2)
\end{aligned}
\end{equation}

$MINVar(V)$ is the variance of the vector $V'$, whose maximum element is the same with the $V$'s and other elements have the same value $\frac{1-max(V)}{C-1}$. $MINVar(V)$ is the smallest variance among vectors whose maximum are same with $V$'s, so that $\frac{MINVar(V)}{Var(V)}$ measures the concentration of the probabilities distribution. According to Eq.~\ref{eq4}, we can prove that the uncertainty score has three properties: (1) it has a boundary of [0,1); (2) it has a negative correlation with the value of maximum probability; (3) it has a positive correlation with the concentration of the probabilities distribution. Due to these properties, the uncertainty score
has a good response to the informativeness of samples.

The objective function of the discriminator is defined as follows,

\begin{equation}\label{eq5}
\begin{aligned}
\mathcal{L}^{D} = &- E[log (D(q_\theta (z_L \mid x_L)))]  \\
 &- E[log (Indicator(x_U) - D(q_\theta (z_U \mid x_U)))]
\end{aligned}
\end{equation}
where the indicator relabels the unlabeled data's label.

As adversarial learning, the objective function of the unified representation generator in SRAAL is
\begin{equation}\label{eq6}
\begin{aligned}
\mathcal{L}^{G}_{adv} = &- E[log (D(q_\theta (z_L \mid x_L)))] \\
&- E[log (D(q_\theta (z_U \mid x_U)))]
\end{aligned}
\end{equation}

The total objective function combined with Eq.~\ref{eq1}, Eq.~\ref{eq3} and Eq.~\ref{eq6} for the latent variable generator is also given as follows,
\begin{equation}\label{eq7}
\begin{aligned}
\mathcal{L}^{G} = \lambda_1 \mathcal{L}^{UIR} + \lambda_2 \mathcal{L}_{L}^{STL} + \lambda_3 \mathcal{L}^{G}_{adv}
\end{aligned}
\end{equation}

\subsection{Sampling strategy in active learning}
\label{3.4}

The algorithm for training the SRAAL at each iteration is shown in Fig.~\ref{img2}. In the sampling step, the generator generates the unified representation for each unlabeled sample. The discriminator predicts its state value, and the top-K samples are selected to be labeled by the oracle.
\subsection{Initially sampling algorithm}
\label{3.5}

\renewcommand{\algorithmicrequire}{\textbf{Input:}}
\renewcommand{\algorithmicensure}{\textbf{Hyperparameters:}}
\begin{algorithm}[t]
\setstretch{1.1}
\caption{Initialization of labeled pool}
\label{alg:B}
\begin{algorithmic}
\REQUIRE {labeled data pool $D_L$, unlabeled pool $D_U$, the size of initial labeled pool $\mathcal{M}$, latent variables $z$ for all the data points}
\ENSURE {Randomly select $I$ ($I \ll \mathcal{M}$) data points in  $D_U$ and move them to  $D_L$}
\REPEAT
\STATE $u = argmax_{x_U \in D_{U}} [min_{x_L \in D_{L}} Distance(x_U,x_L)]$
\STATE $D_L = D_L $ $\cup$   \{$u$\}
\STATE $D_U = D_U $ $ - $ $  $\{$u$\}

\UNTIL{ $size(D_L) = \mathcal{M}$ }
\RETURN the initialized labeled pool $D_L$
\end{algorithmic}
\label{a1}
\end{algorithm}

It is worth noting that most AL methods mainly study the selection strategy, while the initialization of labeled pool is usually random. However, the initialization of labeled pool can heavily affect subsequent sample selection and performance of active learning. Thus, we propose an initially sampling algorithm in which the problem of initially sampling is defined as a set cover problem. The goal cover problem is to find a subset of data points that the largest distance of any point to the subset is minimum. To measure the distance between samples, first we train the unsupervised image reconstructor so that it learns the latent variables of all the samples, then we apply a greedy k-center algorithm where the distance between two samples is measured by the Euclidean distance between their latent variables. The final output is a subset with $\mathcal{M}$ samples that is labeled by an oracle and sent to the labeled pool. Alg.~\ref{a1} shows the detail of the algorithm.

\section{Experiment}

\begin{figure*}[t]
\centering
\subfigure{
	\begin{minipage}{0.49\linewidth}
	\centering
	\includegraphics[width=\linewidth,trim=40 200 40 228]{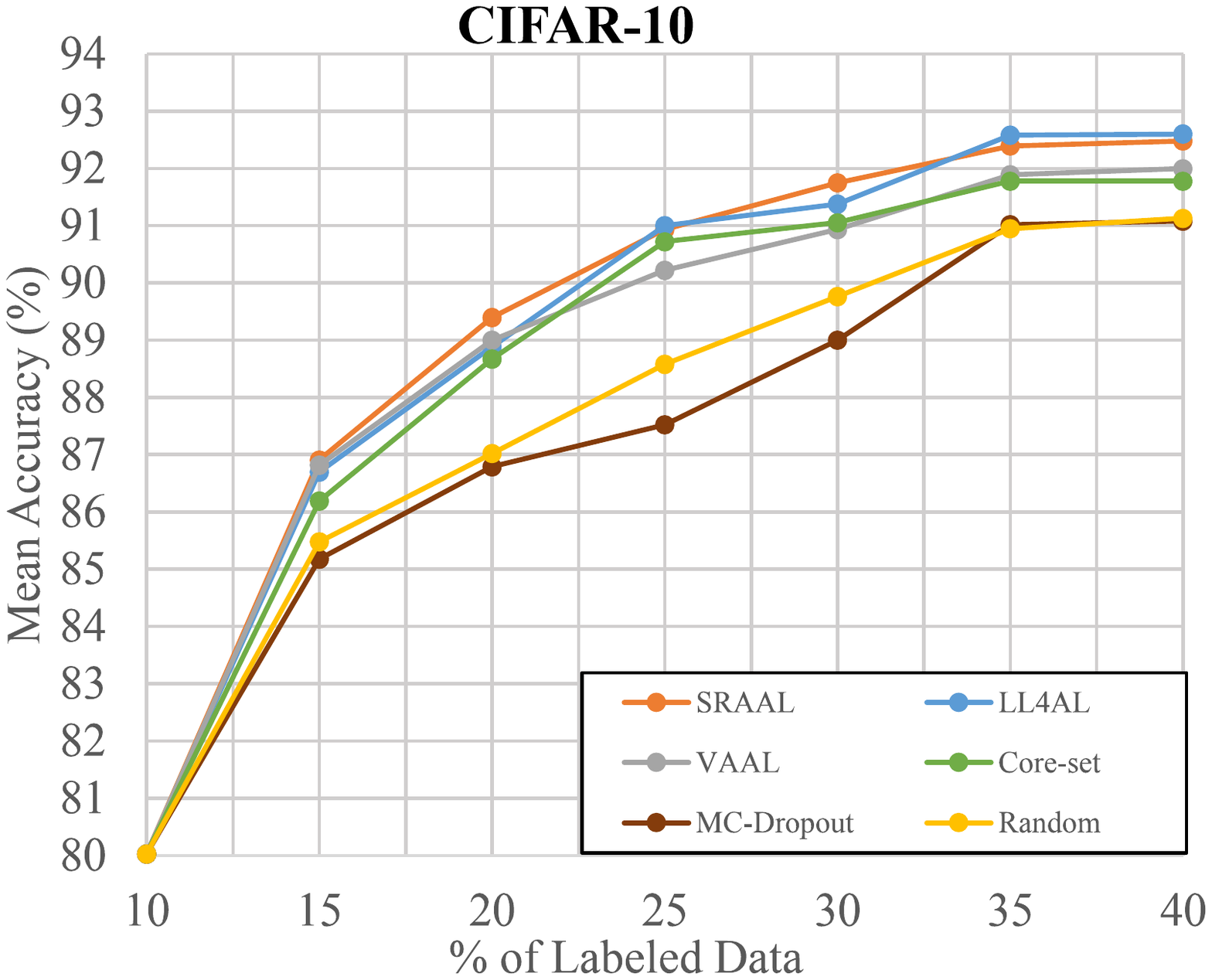}
	\end{minipage}
}
\subfigure{
	\begin{minipage}{0.47\linewidth}
	\centering
	\includegraphics[width=\linewidth,trim=100 250 100 225]{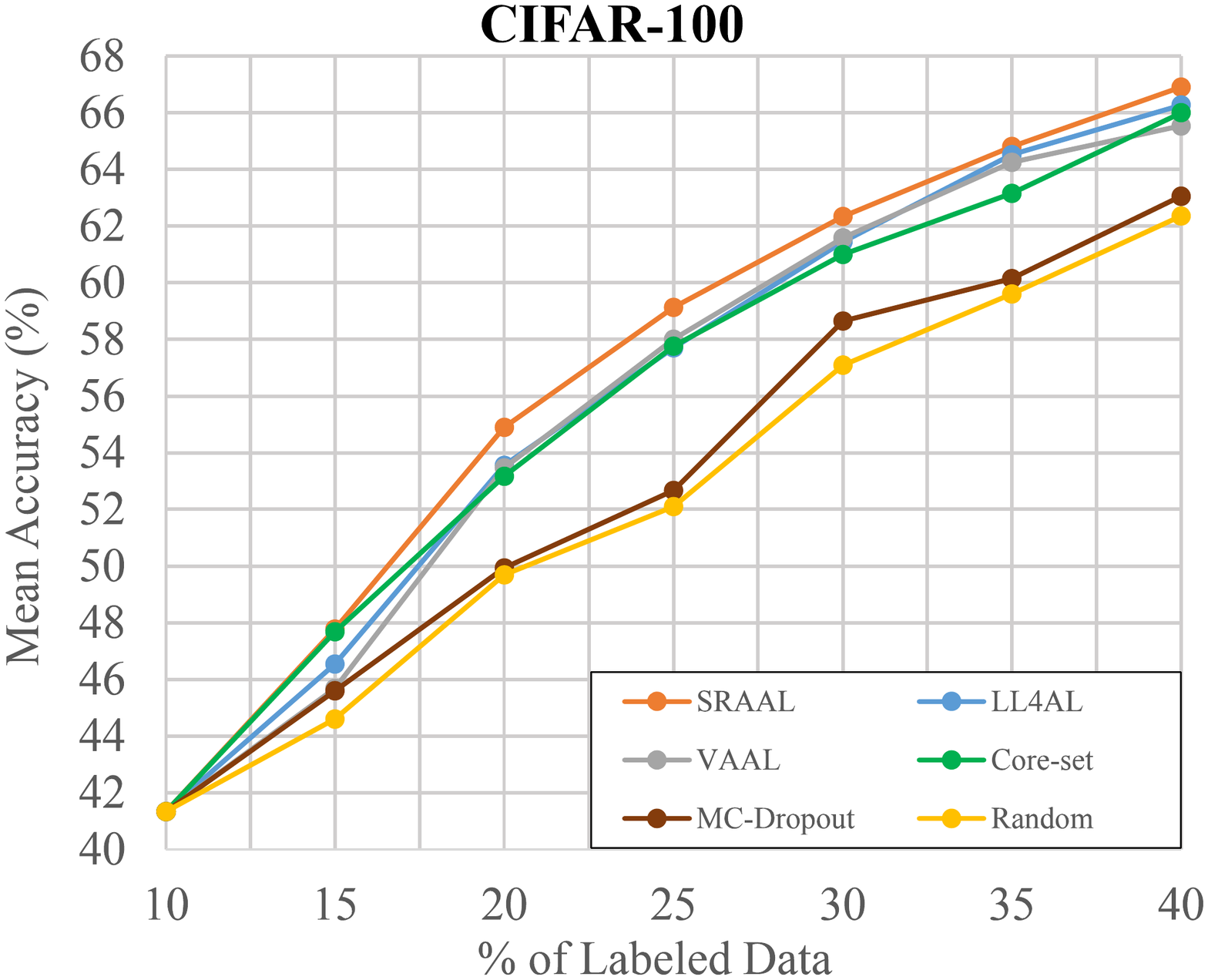}
	\end{minipage}
}
\vspace{-0.2cm}
\caption{Active learning results of image classification over CIFAR-10 and CIFAR-100.}
\vspace{-0.4cm}
\label{f1}
\end{figure*}

In this section, we evaluate SRAAL against state-of-the-art active learning approaches on image classification and segmentation task.

For both tasks, we initialize the labeled pool $D_L^0$ by randomly sampling M = 10\% samples from the entire dataset and the rest 90\% samples make up the initial unlabeled pool $D_U^0$. The unlabeled pool contains the rest of the training set form which samples are selected to be annotated by the oracle. We iteratively train the current model and select K = 5\% samples from the unlabeled pool until the portion of labeled samples reaches 40\%. For each active learning method, we
repeat the experiment 5 times with different initial labeled pool and report the mean performance. When we compare the performance with our methods, they both start with the same initial labeled pool.

To verify the performance of our initial sampling algorithm, we also set an experiment to compare the designed
initialization with the random style. Besides these experiments, we also set an experiment for ablation study to evaluate some main modules in our model.

\subsection{Active learning for image classification}
\label{4.1}

\noindent \textbf{Dataset.} For the classification task, We choose CIFAR-10, CIFAR-100~\cite{krizhevsky2009learning} and Caltech-101~\cite{FeiFei2004LearningGV} as they are classical for image recognition and some recent works evaluate on them. Both of CIFAR-10 and CIFAR-100 have 60,000 images of 32$\times$32$\times$3 where 50000 is training images and 10000 is test images. The CIFAR-10 with 10 categories has 6000 images per class, while the CIFAR-100 has 100 classes containing 600 images each. The Caltech-101 consists of a total of 9,146 images, split between 101 different categories, as well as a background category, and each category in Caltech-101 has about 40 to 800 images. These datasets simulate different real-world situations for the number of images per class.

\noindent \textbf{Compared methods.} For image classification,we compare the performance of SRAAL with some recent state-of-the-art approaches, including Core-set~\cite{sener2017active}, Monte-Carlo Dropout~\cite{gal2016dropout}, VAAL~\cite{sinha2019variational} and LL4AL. We also introduce the random selection method as a baseline. We reproduce the results of these works with the official released
codes and adopt the original hyperparameters. When evaluating, these methods are evaluated by the same target
model.

\noindent \textbf{Performance measurement.} We evaluate the performances of these methods in image classification by measuring the average accuracy of 5 trials. In each trial, all the methods begin with a same initial labeled pool.  
The target model used to evaluate the accuracy is a 18-layer residual network (ResNet-18). We utilize a specified Resnet-18 model for CIFAR-10/100 and a classical Resnet-18 for Caltech-101. Besides, images from Caltech-101 are resized to 220$\times$220 for convenience.

\subsubsection{Performance on CIFAR-10}

The left of Fig.~\ref{f1} shows the performances on CIFAR-10. 
We can observe that, first, our method achieves an accuracy over 90\% by using 25\% of the data and the performance in last iteration reaches 92.48\%. The highest accuracy of the Resnet-18 with full dataset reaches 93.5\%, which is only 1.02\% better than SRAAL with 40\% samples. 
This shows that on CIFAR-10 SRAAL performs closely to the full-data trained model. 
Second, our method evidently outperforms MC-Dropout, Random sample, core-set and VAAL and is on par with the LL4AL. LL4AL outperforms our SRAAL at 25\%, 35\% and 40\% with very slight lead, but underperforms our method at 15\%, 20\% and 30\%. This also demonstrates that our SRAAL has selected more informative samples, and it benefits from the use of annotation and labeled/unlabeled state information.

\subsubsection{Performance on CIFAR-100}
CIFAR-10 dataset has 50000 training images categorized into 10 classes, while the CIFAR-100 has 50000 images in 100 classes. 
Thus, this dataset is much more challenging. 
The right of Fig.~\ref{f1} shows the performances, we find that, first, all the AL methods have better results than random selection method. 
Second, on CIFAR-10, LL4AL performs continuously better than most methods. 
However, the LL4AL on CIFAR-10 becomes not as competitive as on CIFAR-10. 
Especially in first iteration, the coreset and LL4AL achieve better performance than LL4AL.
The LL4AL trains its model only with the labeled samples.
The inadequate labeled samples restrain the accuracy of its main module, which makes the sample selection inefficient. Third, for all iterations, our SRAAL achieves the better performance than the state-of-the-art methods, such as VAAL, LL4AL. 
Although the VAAL also uses the label state information, the state relabeling in the discriminator from our SRAAL has a better use of it. Besides, the annotation embedded unified representation from the generator provides the richer feature for samples.

\subsubsection{Performance on Caltech-101}
\begin{figure}[t]
\begin{center}
\includegraphics[width=0.91\linewidth,trim=80 250 80 230]{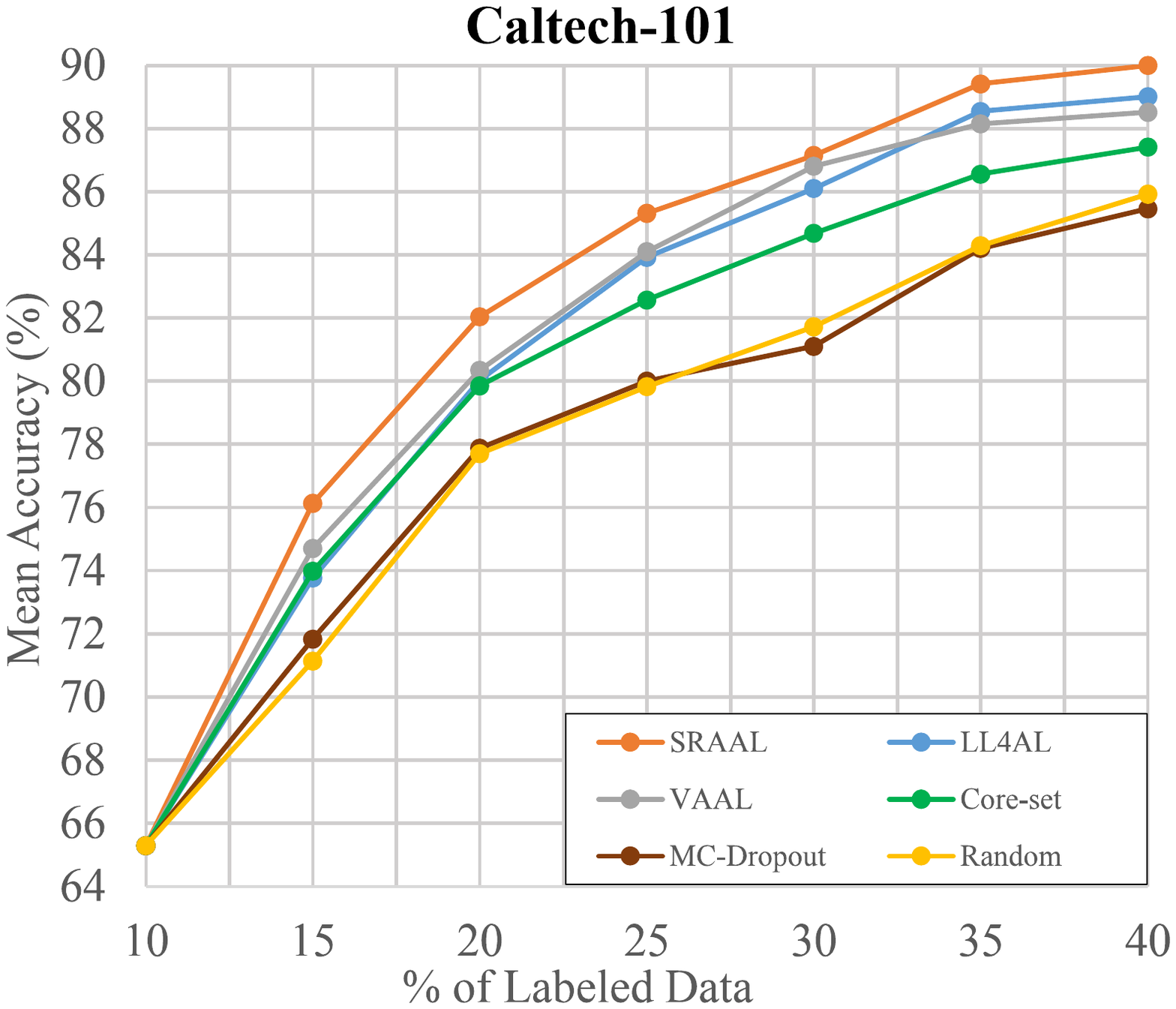}{\rule{0pt}{2in} \rule{1\linewidth}{0pt}}
\end{center}
   \caption{Active learning results of image classification over Caltech-101.}
\label{f2}
\end{figure}

To further explore the influence of image amount per class to AL model, we conduct the comparison experiment on the Caltech-101 dataset.
Fig.~\ref{f2} shows the performance of these methods. 
Caltech-101 has less images per class and the image amount for each class is also different. 
We find that SRAAL outperforms all previous methods from the first iteration to last, and the gap between SRAAL and second-best method becomes larger than that over CIFAR100. 
This phenomenon provides a further proof that our method can better resist the impact from adequate labeled samples. 
Besides, the core-set method and LL4AL method perform worse than VAAL because they only utilize annotation information. 
This verifies that the label state information is useful to help sample the representative data again.

\subsection{Active learning for semantic segmentation}
\begin{figure}[t]
\begin{center}
\includegraphics[width=1\linewidth,trim=140 120 120 40]{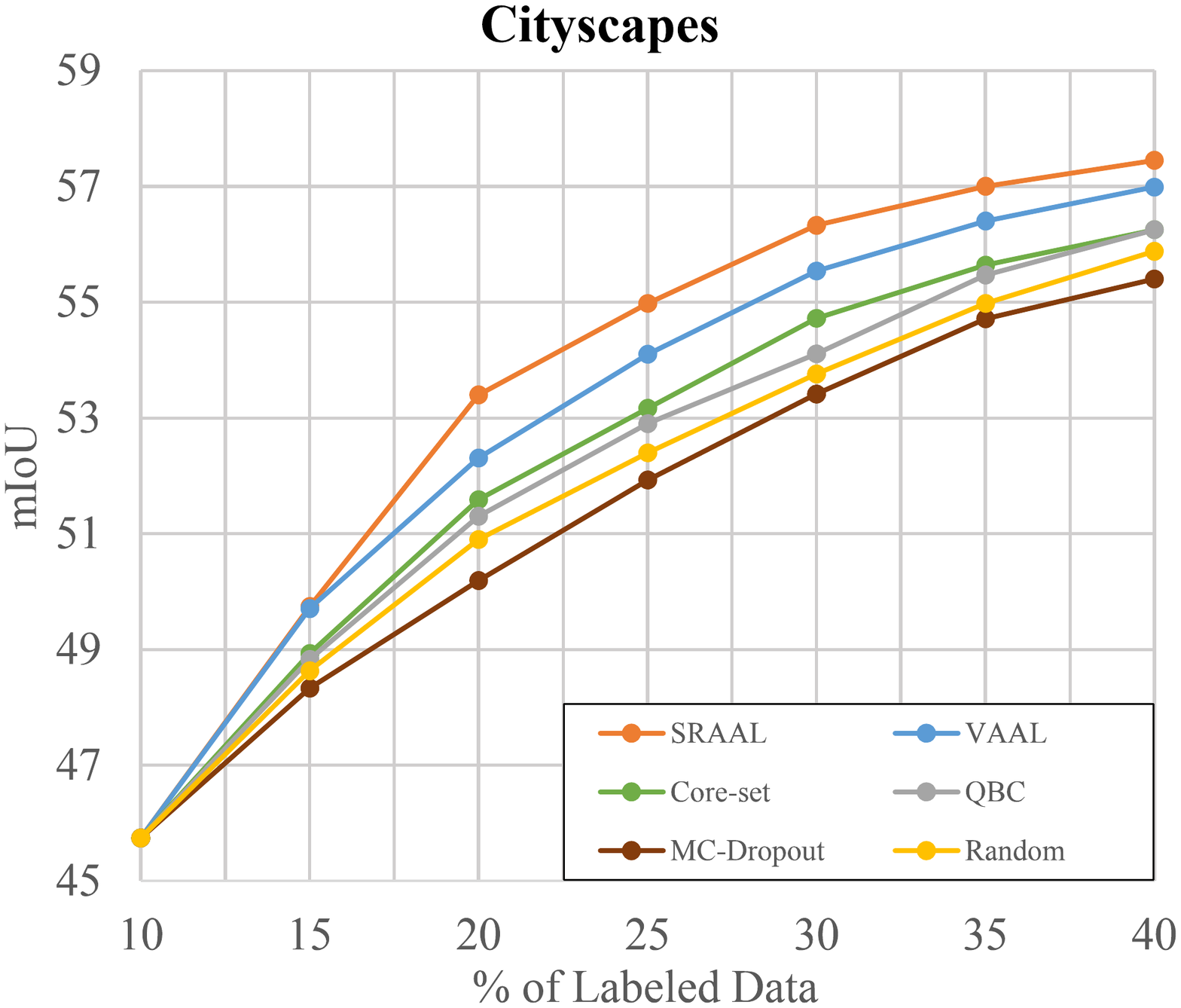}{\rule{0pt}{2in} \rule{1\linewidth}{0pt}}
\end{center}
   \caption{Active learning results of image semantic segmentation over Cityscapes.}
\label{f3}
\end{figure}
\noindent \textbf{Dataset.} For semantic segmentation, it is a popular task to evaluate the active learning model. 
Semantic segmentation is more challenging than image classification, so that this experiment evaluates the performance of AL methods on a
difficult task. 
Here we also conduct the comparison experiment on the Cityscapes dataset~\cite{cordts2016Cityscapes}. 
This dataset has 3475 frames with instance segmentation annotations recorded in street scenes. Following \cite{sinha2019variational}, we convert this dataset into 19 classes.

\noindent \textbf{Compared methods.} We evaluate our SRAAL against some active learning approaches for semantic segmentation. The compared works include Core-set~\cite{sener2017active}, MC-Dropout~\cite{gal2016dropout}, Query-By-Committee (QBC)~\cite{KuoCost}, suggestive annotation (SA)~\cite{yang2017suggestive} and VAAL~\cite{sinha2019variational}.

\noindent \textbf{Performance measurement.} For this task, the target model is DRN, and the mean IoU is used to evaluate the performances. 
All the methods are evaluated with a same initial labeled pool and a same selection budget for each iteration.

Fig.~\ref{f3} shows our results of semantic segmentation about
different methods. 
We can observe that, first, the VAAL and our SRAAL obtain better performance than other methods, such as SA, QBC, MC-Dropout. 
This is because both VAAL and SRAAL introduce the label state information to model the sample selection. 
Second, our SRAAL outperforms the VAAL with a large margin. 
This benefits from the state relabeling of the proposed online uncertainty indicator. 
This relabeled state can better guide the discriminator to choose the most informative samples.

\subsection{Initialization algorithm comparison}
\begin{figure}[t]
\begin{center}
\includegraphics[width=1\linewidth,trim=50 270 70 240]{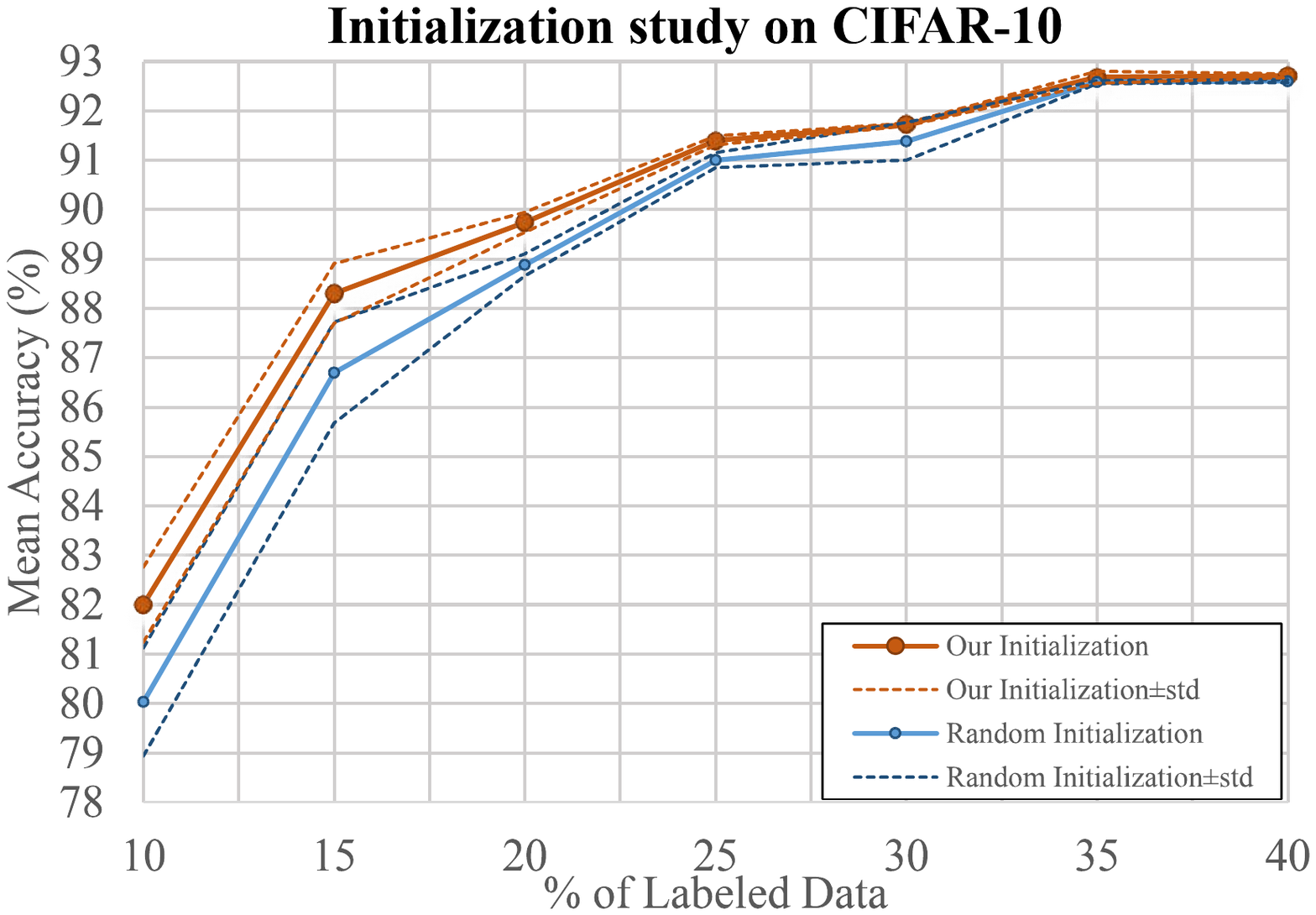}{\rule{0pt}{2in} \rule{1\linewidth}{0pt}}
\end{center}
   \caption{Experiment result for initial sampling algorithm.}
\label{f4}
\end{figure}

As mentioned in Section~\ref{3.5}, we introduce the k-center approach to initialize the labeled pool. 
We evaluate the initialization algorithm on the CIFAR-10 dataset. 
The AL model is our proposed SRAAL and the target model is the ResNet-18 image classifier, which is the same with that in Section~\ref{4.1}.

As shown in Fig.~\ref{f4}, we can observe that the mean accuracy of our initialization algorithm is significantly higher than the random initialization. 
The higher accuracy proves that our initial labeled samples are more informative than random selected ones. 
Further, the dotted lines show that the standard deviation of our initialization is also less than that of the random initialization. To sum up, our initialization makes subsequent sampling more efficient.

\subsection{Ablation study}

To evaluate the contribution of different modules in our model, we conduct this experiment for ablation study on CIFAR-100 dataset. 
As the state relabeling is the key of our work, we first verify the role of it by eliminating the online uncertainty indicator and using the original binary label state. 
We also perform an ablation on the supervised target learner to explore the importance of the annotation embedded unified representation. 
Besides, an ablation to both two modules is performed as the control group.

\begin{figure}[t]
\begin{center}
\includegraphics[width=0.96\linewidth,trim=70 240 70 220]{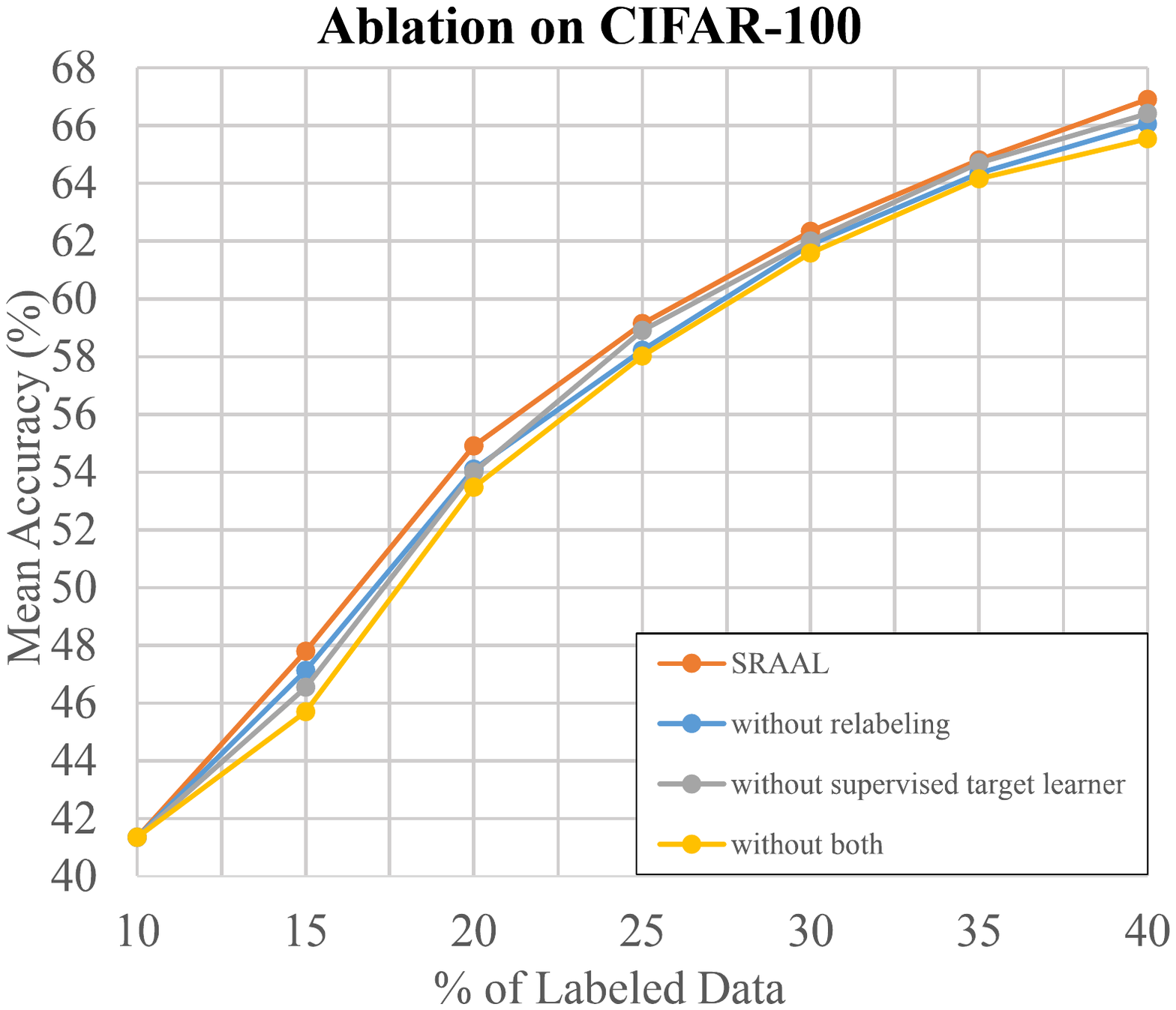}{\rule{0pt}{2in} \rule{1\linewidth}{0pt}}
\end{center}
   \caption{Experiment result for ablation study on CIFAR-100.}
\label{f5}
\end{figure}

Fig.~\ref{f5} shows the results for the ablation study. The complete SRAAL consistently outperforms all the ablations, and the ablation to two modules performs yields lowest accuracy among the ablations. The above phenomenon illustrates that either the relabeling or the annotation-embedded representation helps to improve the AL performance.

\subsection{Comparison on different uncertainty estimators}

\begin{table}[b]
\begin{tabular}{|p{0.95cm}|p{0.8cm}|p{0.8cm}|p{0.8cm}|p{0.8cm}|p{0.8cm}|}
\hline
  (\%)     & 20\%  & 25\%  & 30\%  & 35\%  & 40\%  \\ \hline
Ours       & 55.0  & 59.1  & 62.3  & 65.0  & 66.9  \\ \hline
Entropy    & 54.0  & 58.2  & 61.1  & 64.5  & 65.7  \\ \hline
SD         & 54.1  & 57.0  & 59.4  & 63.3  & 64.1  \\ \hline

\end{tabular}
\vspace{0.1cm}
\caption{Comparison with entropy and standard deviation(SD) under different sampling ratios.}
\label{tab2}
\end{table}

To accurately relabel the state of unlabeled data with different importance, we design an uncertainty score(Eq.~\ref{eq4}) in the online uncertainty indicator module. 
To verify the superiority of our score and prove that our uncertainty score is more suitable for state relabeling, we compare some common uncertainty acquisition functions with ours by replacing our uncertainty score with them.
The experiment result in Tab.~\ref{tab2} shows that our indicator outperforms these uncertainty acquisition functions under different sampling ratios, which verifies that our uncertainty score can better reflect the importance of unlabeled data.



\section{Conclusion}
In this paper, we study the active learning and propose a state-relabeling adversarial active learning model (SRAAL) that makes full use of both the annotation and the state information for deriving most informative unlabeled samples.
The model consists of a unified representation generator that learns the annotation-embedded image feature, and a labeled/unlabeled state discriminator that selects most informative samples with the help of online updated indicator. 
Further, we introduce the k-center approach to initialize the labeled pool, which makes subsequent sampling more efficient. 
The experiments on image classification and segmentation demonstrate that our model outperforms previous state-of-the-art methods and the initially sampling algorithm significantly improve the performance of our model.

\textbf{Acknowledgement.} This work was supported in part by the National Key R\&D Program of China under Grand:2018AAA0102003, in part by National Natural Science Foundation of China: 61771457, 61732007, 61772497, 61772494,  61931008, 61620106009, U1636214, 61622211, U19B2038, and in part by Key Research Program of Frontier Sciences, CAS: QYZDJ-SSW-SYS013 and the Fundamental Research Funds for the Central Universities under Grant WK2100100030.

{\small
\bibliographystyle{ieee_fullname}
\bibliography{egbib}
}

\end{document}